%% file: main.tex
\documentclass{article}
\usepackage{tikz}
\usepackage{textcomp}

\usepackage{lipsum}

\usepackage{spconf,amsmath}
\ninept
\usepackage{xspace}
\usepackage[position=top]{subfig}  %
\usepackage{booktabs} %
\usepackage{url} %

\usepackage{float}
\floatstyle{plaintop}
\restylefloat{table}

\makeatletter
\def\endthebibliography{%
	\def\@noitemerr{\@latex@warning{Empty `thebibliography' environment}}%
	\endlist
}
\DeclareRobustCommand\onedot{\futurelet\@let@token\@onedot}
\def\@onedot{\ifx\@let@token.\else.\null\fi\xspace}

\makeatother

\usepackage{graphicx}
\usepackage{xcolor}

\newif\ifSuppressMemo
\ifSuppressMemo
\newcommand{\memo}[1]{}
\else
\usepackage{color}
\newcommand{\memo}[1]{{\bf \textcolor{red}{[#1]}}}
\fi
\title{MULTI-KERNEL PREDICTION NETWORKS FOR DENOISING OF BURST IMAGES}
\name{Talmaj Marin\v{c}, Vignesh Srinivasan, Serhan G\"{u}l, Cornelius Hellge, Wojciech Samek
\thanks{This research has received funding from the German Federal Ministry for Economic Affairs and Energy under the VIRTUOSE-DE project and by the German  Federal Ministry for Education through the Berlin Big Data Center under Grant 01IS14013A and the Berlin Center for Machine Learning under Grant 01IS18037I.}
\thanks{\copyright 2019 IEEE. Personal use of this material is permitted. Permission from IEEE must be obtained for all other uses, in any current or future media, including reprinting/republishing this material for advertising or promotional purposes, creating new collective works, for resale or redistribution to servers or lists, or reuse of any copyrighted component of this work in other works.}
}
\address{Fraunhofer Heinrich Hertz Institute, Berlin, Germany}

\begin{document}

\maketitle

\begin{abstract}
In low light or short-exposure photography the image is often corrupted by noise. While longer exposure helps reduce the noise, it can produce blurry results due to the object and camera motion.
The reconstruction of a noise-less image is an ill posed problem.  
Recent approaches for image denoising aim to predict kernels which are convolved with a set of successively taken images (burst) to obtain a clear image.
We propose a deep neural network based approach called Multi-Kernel Prediction Networks (MKPN) for burst image denoising.
MKPN predicts kernels of not just one size but of varying sizes and performs fusion of these different kernels resulting in one kernel per pixel. 
The advantages of our method are two fold: (a) the different sized kernels help in extracting different information from the image which results in better reconstruction and (b) kernel fusion assures retaining of the extracted information while maintaining computational efficiency. %
Experimental results reveal that MKPN outperforms state-of-the-art on our synthetic datasets with different noise levels.

\end{abstract}
\section{Introduction}
\label{sec:intro}

\input{related_work}

\section{Proposed method}
\label{sec:model}
\input{method.tex}

\section{Experiments}
\label{sec:typestyle}
\input{results.tex}
\section{Conclusion}
\label{sec:conclusion}
Burst image denoising, with its inherent challenges, remains to be an open problem. 
In this work, we propose MKPN, a DNN based method for denoising of burst images captured by handheld cameras. 
The novelty of this method lies in predicting kernels of different sizes and performing kernel fusion by in-place addition before the convolution operation. 
MKPN effectively combines the best behavior of small and large kernels -- it manages to denoise flat areas as well as preserve the detailed image structures. 
Kernel fusion ensures that MKPN is able to extract different information from the different kernels without compromising on computational efficiency. 
We empirically show that MKPN 
outperforms state-of-the-art models quantitatively and provides visually pleasing results. 

\bibliographystyle{IEEEbib}
\bibliography{denoising-bibliography}

\end{document}

%% file: related_work.tex
Image denoising is a long-standing problem finding applications in fields ranging from astronomical imaging to hand-held photography. With the development of digital photography and smartphone technology, it has recently become possible to take high-quality photos using relatively inexpensive equipment. However, there are still major differences between the imaging capabilities of hand-held devices such as smartphones and professional equipment such as DSLRs. One of the most important factors for taking noise-free photos is to collect as much light as possible. Professional cameras contain several hardware solutions such as large aperture lenses, sensors with large photosites and high-quality A/D converters for increased light collection~\cite{godard2018deep}. However, for the sake of compactness, smartphones contain smaller and less expensive variants of these hardware elements which can result in noisy imaging, especially in low light conditions. To combat this problem, image denoising algorithms are implemented as a software solution in most smartphones~\cite{hdrplus}. %

Classical methods for image denoising developed in the early 1990s like antisotropic diffusion~\cite{perona1990scale} and total variation denoising~\cite{rudin1992nonlinear} use analytical priors and non-linear optimization to recover a clear image.
More recently, plain neural networks~\cite{burger2012image}, convolutional neural networks~\cite{jain2009natural} and neural networks with auto-encoder architectures were used for single image denoising~\cite{vincent2010stacked, xie2012image, agostinelli2013adaptive}. 
The combination of the last two, a convolutional auto-encoder, showed promising results in medical image denoising~\cite{gondara2016medical}.
The same architecture was further improved by addition of skip connections~\cite{resnet} placed between the encoder and the decoder. This architecture was used both in single image denoising~\cite{mao2016image, brooks2018unprocessing} as well as in denoising of the images captured in quick succession, i.e. \emph{burst} image denoising~\cite{mildenhall2018burst}. 

Burst image denoising methods operate on a set of successive, rapidly taken images to compute a single, noise-free result. Since the noise is usually randomly distributed and a set of images obtained by the same camera have similar characteristics, it is reasonable to expect that burst image denoising tends to work better than single image denoising in most cases.
Burst denoising methods commonly first align the successive images as a pre-processing step, and then fuse and denoise the aligned images~\cite{liu2014fast, hdrplus, godard2018deep, Chen2018LearningTS}. 
Many of the state-of-the-art approaches in burst image denoising are based on fully convolutional neural network architectures~\cite{godard2018deep, mildenhall2018burst, Chen2018LearningTS, kokkinos2018iterative}.

Neural network based image denoising methods commonly operate on the whole image and predict the denoised image directly. Another approach is to design a network that can learn to predict spatially variant kernels for each pixel in the input. These \emph{per-pixel} kernels are then convolved over the input to produce the final output. 
This approach worked very well in video frame interpolation~\cite{frame_interpolation}, denoising of renderings~\cite{Bako17, vogels2018denoising} and burst image denoising~\cite{mildenhall2018burst}.

In this paper we propose a novel neural network based method burst image denoising method. Our method is conceptually similar to~\cite{mildenhall2018burst} but it operates in a fundamentally different way. Specifically, instead of predicting a single, fixed-size kernel for each pixel, our method predicts kernels of multiple sizes for each pixel taking into account the spatially varying image structures. This allows our method to adapt to the properties of image structures with different properties (e.g. textured vs. non-textured, homogeneous areas) and successfully denoise a wide range of images without having to manually tweak the kernel size depending on the characteristics of the image.

%% file: method.tex
\subsection{Multi-Kernel Prediction Network}
\label{sec:mkpn}

\begin{figure*}[t]
\begin{center}
   \includegraphics[width=0.76\linewidth]{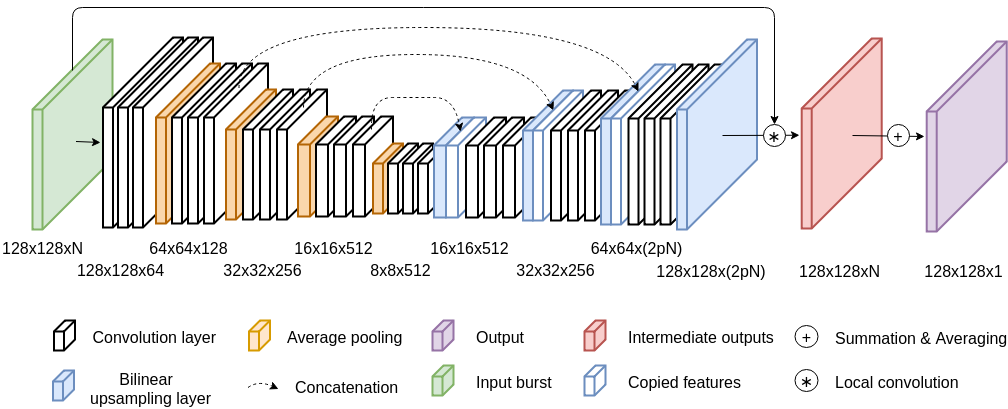}
\end{center}
   \caption{The proposed Multi-Kernel Prediction Network architecture. It predicts per pixel kernels of various sizes for each of the input images in a burst. The input of the network is a burst sequence of length $N=8$, $p$ is the sum of predefined kernel sizes and $S$ is the set of these kernels. Each of the burst images is deconvolved by the predicted per pixel kernels as seen in Fig. \ref{fig:per_pixel_convolution} and averaged to the final output. The numbers below the blocks represent the 3-dimensional data structure at different levels of the architecture. The dimensions correspond to the $\textit{height}\times \textit{width}\times \textit{channel}$.}
\label{fig:architecture}
\end{figure*}

Our goal is to perform denoising on burst images corrupted by noise due to low light or shot exposure photography. Given a noisy set of input burst images, kernels are predicted for each pixel using a deep neural network. These kernels are then convolved with their respective input pixels in the respective burst image which are then averaged to reconstruct the denoised image.~\cite{mildenhall2018burst} performed this denoising by predicting kernels of a predetermined size as given in \eqref{eq:kpn}:
\begin{align} \label{eq:kpn}
    \hat{I}(x,y) &= \frac{1}{N} \sum_{i=1}^{N} K_{i}(x,y) * P_{i}(x,y).
\end{align}
However, images consist of heterogeneous patches where every pixel in a patch is different compared to its surrounding pixels.  
Flat regions like sky in an image would provide for a more certain prediction while densely cluttered leaves in an image would provide for more uncertainties in the prediction. The predicted kernels should be able accommodate to the different regions and pixels in the image. Naturally, kernels of predetermined size cannot successfully adapt to the diversity of pixels in the images.

Our method, Multi-Kernel Prediction Networks (MKPN) is a direct extension of the~\cite{mildenhall2018burst} and~\cite{frame_interpolation}. To the best of our knowledge this is the first attempt to utilize multiple kernels of different sizes predicted by a neural network for image denoising. Although~\cite{mai2015kernel} combine different kernels, they assume a fixed size of the kernel and do not use a deep learning based approach. 
MKPN predicts kernels of different sizes for each pixel in the image belonging to the set of burst images instead of predicting fixed-size kernels. The predicted kernels of different sizes are then convolved with each pixel in the input image from the set of noisy burst images which are then averaged to obtain the final reconstruction as described in \eqref{eq:mkpn}:
\begin{align} \label{eq:mkpn}
    \hat{I}(x,y) &= \frac{1}{N \times |S|} \sum_{i=1}^{N} \sum_{s\in S} K_{i}^{s}(x,y) * P_{i}^{s}(x,y).
\end{align}
Here $N$ is the length of the input burst, $S$ is the set of kernel sizes, $K_{i}^{s}(x,y)$ is a kernel with size $s$ for pixel located at $I_i(x,y)$ and $P_{i}^{s}(x,y)$ is a patch with size $s$ in $I_i$ centered at $(x,y)$. $I_i$ is the $i$-th image from the input burst. %
Kernels of different sizes extract and accumulate information from image structures with different characteristics.

Ideally, MKPN will work well when many kernels of different sizes are used. However, convolving each kernel with its corresponding patch significantly increases the required amount of computations. In order to reduce the amount of computations, we apply two enhancements:

    \textbf{Separable Kernel Estimation}: We approximate the 2D kernels by a linear combination of separable 1D kernels~\cite{Rigamonti2013LearningSF, frame_interpolation}. In this way the number of learnable parameters is reduced from $n^2$ to $2n$ for each kernel of size $n\times n$, significantly reducing the computation cost.
    
    \textbf{Kernel Fusion}: Instead of convolving each kernel separately with each image in the burst, MKPN performs in-place kernel addition as described in \eqref{eq:inplace1}, before the convolution operation: 
    \begin{align} \label{eq:inplace1}
        \Tilde{K}_i(x,y) &= \frac{1}{|S|} \sum_{s\in S} K_{i}^{s}(x,y).
    \end{align}
    The accumulated kernels $\Tilde{K}_i(x,y)$ are then convolved with the corresponding image patches in a single operation, as shown in \eqref{eq:inplace2}: 
    \begin{align} \label{eq:inplace2}
        \hat{I}(x,y) &= \frac{1}{N} \sum_{i=1}^{N} \Tilde{K}_i(x,y) * P_{i}(x,y).
    \end{align}
    This essentially brings the number of convolution operations equal to that of~\cite{mildenhall2018burst}. The computational cost of the in-place additions is negligible. Note that model compression \cite{han2015deep, WieArXiv18b} and efficient representations \cite{WieArXiv18} can further reduce the computational costs.

\subsection{Network Architecture}
Fig.~\ref{fig:architecture} shows an overview of the MKPN architecture. MKPN has a typical U-Net~\cite{unet} shape resembling the architectures of~\cite{mildenhall2018burst, frame_interpolation}. The network consists of convolutional layers, ReLU %
activation functions, average pooling layers and billinear upsampling layers. The convolutional layers use $3 \times 3$ filters with zero padding and stride one, and are always followed by a ReLU activation function. The average pooling layers have a pool size of $2 \times 2$ and stride two, which in effect decreases the spatial resolution by a factor of two. On the contrary, the billinear upsampling layer increases the spatial resolution by a factor of two. 

A series of convolutional and average pooling layers encode the input features into a latent representation. Series of bilinear upsampling and convolutional layers decode these features to predict the per pixel kernels. After each upsampling layer, the high resolution features from the encoder side of the architecture are concatenated to the decoder side.

\begin{figure}[t]
\begin{center}
   \includegraphics[width=0.8\linewidth]{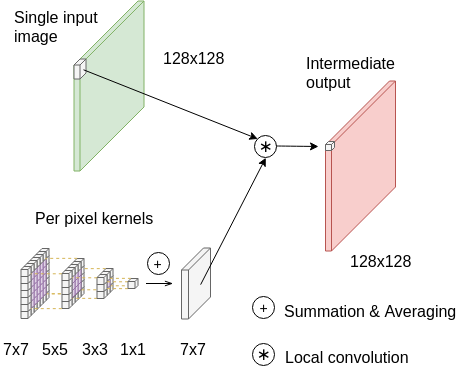}
\end{center}
   \caption{Kernel fusion of the kernels of various sizes. For each kernel size $128^2$ kernels are predicted. Each kernel corresponds to a pixel in the input image. Kernels corresponding to the same pixel are averaged at inference. Convolving all kernels over a single input image results in an intermediate output, which is a part of intermediate outputs in Fig.~\ref{fig:architecture}.}
\label{fig:per_pixel_convolution}
\end{figure}

The channel dimension of the layers of the last convolution block is $2p\cdot N$, where $p$ is the sum of the predefined sizes of different kernels and $N$ is the number of images in the input burst. For example, if the selected kernel sizes are $5 \times 5$ and $11 \times 11$, then $p = 5 + 11 = 16$. 
In addition, another billinear upsampling layer is employed which scales up the learned linear combinations of 1D per pixel kernels of various sizes to match the input shape. Outer products of 1D kernels are computed to produce the 2D kernels. These kernels are applied on the input burst using local convolution as shown in Fig.~\ref{fig:per_pixel_convolution}. Lastly, the deconvolved burst images are averaged to produce the final output. %

\subsection{Training Parameters}
All experimental models were trained in end-to-end fashion using backpropagation. %

Total loss consists of basic and annealing loss as devised in~\cite{mildenhall2018burst}.
Basic loss is composed of mean squared error on pixel intensities and L1 loss on gradient intensities between the denoised image $\hat{I}$ and the ground truth $I$: 
\begin{align}
    \ell(\hat{I}, I) = \lambda_1 \left\Vert \hat{I} - I \right\Vert_2^2 + \lambda_2  \left\Vert \nabla \hat{I} - \nabla I \right\Vert_1.
\label{eq:basic_loss}
\end{align}
The basic loss tries to make the average of all estimations $\hat{I}_i^s$ close to the ground truth $I$.  However, using only the basic loss can lead to convergence at an undesirable local minima that does not utilize all the images in the burst effectively~\cite{mildenhall2018burst}. Therefore, we add a second loss term \emph{annealing loss} to \eqref{eq:basic_loss} that attempts to make each estimation $\hat{I}_i^s$ close to the ground truth $I$ independently. The total loss is obtained as follows:
\begin{align}
    \mathcal{L}(\hat{I}, I) = \ell(\hat{I}, I ) 
                                 + \beta \alpha^t \sum_{i=1}^{N}\sum_{s\in S}  \ell(\hat{I}_i^s, I),
\label{eq:total_loss}
\end{align}
where $N$ is the number of images in the burst, $S$ is the set of kernel sizes, $\beta$ and $\alpha$ are the hyperparameters controlling the weight decay, $t$ is the training step, and $\lambda_1 + \lambda_2 = 1$. 
Please note that during training we discard in-place addition of kernels to enable better convergence. However, once the network is well trained, the in-place addition can help speed up inference.

%% file: results.tex
We followed the same procedure for dataset generation and noise estimation as in~\cite{mildenhall2018burst}.
We trained our models and evaluated their performance on synthetically generated datasets from the Open Images dataset~\cite{open_images}.

\subsection{Data Generation \label{sec:data_generation_denoising}}
The images from the Open Images dataset were $4\times$ downsampled in each dimension using a box filter to reduce noise and compression artifacts. Random patches of size $128\times 128$ were sampled from the images and those were used for both creating the ground truth and the remaining $N - 1$ burst images, where $N$ is the total number of images in the burst. These burst images are offset from the first image by $x_i$ and $y_i$, 
where $x_i$ and $y_i$ are the offsets of image $i$ in horizontal and vertical directions, respectively. The offsets simulate misalignments between consecutive frames caused by hand movements that may occur during handheld photography.
Values for $(x_i, y_i)$ are sampled with probability ${n}/{N}$ from a 2D uniform integer distribution between $[-16,16]$, otherwise from a 2D uniform integer distribution between $[-2,2]$, where $n \sim \mathrm{Poisson}(\lambda)$. 
{The burst images are also considered to be noisy, and hence a signal dependent Gaussian noise is added to the burst:
\begin{equation} \label{eq:noise_distr}
    x_p \sim  \mathcal{N} (y_p, \sigma^2_r + \sigma_s y_p),
\end{equation}
where $x_p$ is the noisy measurement of true intensity $y_p$ at pixel $p$.
Read and shot noise parameters $\sigma_r$ and $\sigma_s$ are sampled uniformly from $[10^{-3}, 10^{-1.5}]$ and $[10^{-2}, 10^{-1}]$, respectively.
}
These ranges were selected from the real observed data. %
Synthetic train dataset is generated on the fly, while the test datasets are pre-generated using different gains (noise levels) that correspond to a fixed set of read and shot noise parameters. The selected values simulate the light sensitivities that correspond to the ISO settings on a real camera. 
The read and shot noise for each gain is as given:
\begin{itemize}
    \item Gain~$\propto1$: $\sigma_r=10^{-2.1}$, $\sigma_s=10^{-2.6}$,
    \item Gain~$\propto2$: $\sigma_r=10^{-1.8}$, $\sigma_s=10^{-2.3}$,
    \item Gain~$\propto4$: $\sigma_r=10^{-1.4}$, $\sigma_s=10^{-1.9}$,
    \item Gain~$\propto8$: $\sigma_r=10^{-1.1}$, $\sigma_s=10^{-1.5}$.
\end{itemize}

\subsection{Noise Estimation}
The camera noise is estimated from the first image in a burst. The noise estimate helps the model denoise beyond the noise levels of the training data \cite{mildenhall2018burst}. It is defined as:
\begin{equation} \label{eq:noise_estimation}
    \hat{\sigma}_p = \sqrt{\sigma^2_r + \sigma_s \max(x_p,0)},
\end{equation}
where $x_p$ is the intensity of pixel $p$ in the first image of a burst. In real data the noise parameters $\sigma_r$ and $\sigma_s$ are available in the DNG raw image format~\cite{dng}. This noise estimate is of the same dimension as the burst images and is appended to the end of the burst.

In all our experiments, we set the parameters as follows $N=8$, $\lambda = 1.5$, $\beta = 100$ and $\alpha = .9998$.
\subsection{Results and Discussion}  
\label{sec:results_denoising}

In the experiments we performed using MKPN, we define the sizes of the kernels in advance -- $S\in \{1,3,5,7,9,11\}$. The kernel sizes and the number of kernels however, are not limited to these and can be used in any combination to produce the best results for a given dataset. 
The two enhancements are utilized in the end-to-end training of MKPN, as explained in Section~\ref{sec:mkpn}. 
We evaluate the performance of MKPN by comparing it to state-of-the-art models for burst image denoising:
\begin{itemize}
    \setlength\itemsep{0.2em}
    \item KPN~\cite{mildenhall2018burst} is a kernel prediction network that outputs per-pixel kernels of \emph{fixed} size (set here to $5 \times 5$) which are then convolved with the input burst images to obtain the output.
    \item KPN $\text{L}_n$ where $n\in \{1,3,5,7,9,11,13,25\}$ and $\text{L}_{n}$ stands for kernel sizes of $n \times n$. Here, the network architecture is the same as KPN but the network predicts separable kernels~\cite{frame_interpolation} to reduce the computational burden (cf. Section~\ref{sec:mkpn}). In total, we train eight models with different kernel sizes $n$.
\end{itemize}

\begin{figure}[t]
\centering
\resizebox{\columnwidth}{!}{
  \subfloat[][input burst] {\includegraphics[width=0.21\linewidth]{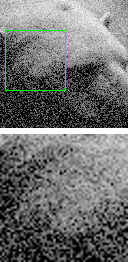}}
  \hspace*{\fill}
  \subfloat[][KPN~\cite{mildenhall2018burst}] {\includegraphics[width=0.21\linewidth]{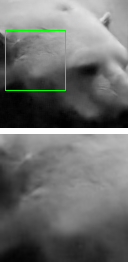}}
  \hspace*{\fill}
  \subfloat[][KPN~$\text{L}_{25}$] {\includegraphics[width=0.21\linewidth]{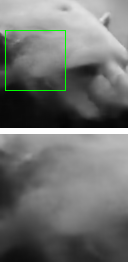}}
  \hspace*{\fill}
  \subfloat[][MKPN] {\includegraphics[width=0.21\linewidth]{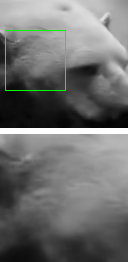}}
  \hspace*{\fill}
  \subfloat[][ground truth] {\includegraphics[width=0.21\linewidth]{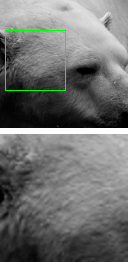}}
}
  \caption{Example of denoising an image of a bear at Gain $\propto$ 4. The detailed fur is recovered best by MKPN. KPN~$\text{L}_{25}$ that uses a large kernel oversmooths the details of the fur. Best viewed on a screen. }
\label{fig:example_bear}
\end{figure}

\begin{figure}[t]
\centering
\resizebox{\columnwidth}{!}{
   \subfloat[][input burst] {\includegraphics[width=0.21\linewidth]{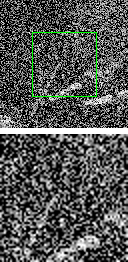}}
   \hspace*{\fill}
   \subfloat[][KPN~\cite{mildenhall2018burst}] {\includegraphics[width=0.21\linewidth]{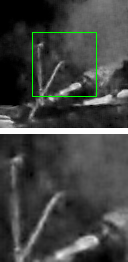}}
   \hspace*{\fill}
   \subfloat[][KPN~$\text{L}_{25}$] {\includegraphics[width=0.21\linewidth]{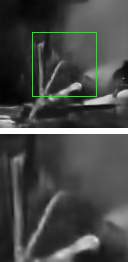}}
   \hspace*{\fill}
   \subfloat[][MKPN] {\includegraphics[width=0.21\linewidth]{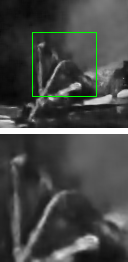}\label{fig:example_grasshoper:mkpn}}
   \hspace*{\fill}
   \subfloat[][ground truth] {\includegraphics[width=0.21\linewidth]{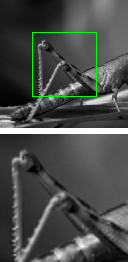}}}
   \caption{Example of denoising an image of a grasshopper at Gain~$\propto$~8. The detailed legs and smooth background are best recovered by the proposed method MKPN. Best viewed on a screen.}
\label{fig:example_grasshoper}
\end{figure}

Each model was trained for 1M iterations and evaluated on four pre-computed test datasets, each one generated with a different noise level. Quantitative analysis is performed using objective quality metrics such as PSNR and SSIM~\cite{ssim} as shown in Table~\ref{table:denoising_exp}. For a qualitative evaluation, we visually inspect the images of MKPN and state-of-the-art models as shown in Fig.~\ref{fig:example_bear} and Fig.~\ref{fig:example_grasshoper}.

\setlength\tabcolsep{3.5 pt}

\begin{table}[t]

\begin{center}
\resizebox{\columnwidth}{!}{
\begin{tabular}{l|rr|rr|rr|rr}
\multicolumn{1}{c}{} & \multicolumn{2}{c}{Gain $\propto$ 1} & \multicolumn{2}{c}{Gain $\propto$ 2} & \multicolumn{2}{c}{Gain $\propto$ 4} & \multicolumn{2}{c}{Gain $\propto$ 8} \\
Model &   PSNR &   SSIM &   PSNR &   SSIM &   PSNR &   SSIM &   PSNR &   SSIM \\
\midrule
MKPN    &  \textbf{35.10} &  \textbf{0.925} &  \textbf{32.22} &  \textbf{0.878} &  \textbf{28.61} &  \textbf{0.781} &  \textbf{25.78} &  \textbf{0.692} \\
KPN~\cite{mildenhall2018burst} &  34.28 &  0.914 &  31.46 &  0.862 &  27.86 &  0.756 &  24.90 &  0.654 \\
KPN~$\text{L}_{25}$ &  34.96 &  0.922 &  32.04 &  0.871 &  28.44 &  0.770 &  25.66 &  0.686 \\
KPN $\text{L}_{13}$ &  34.71 &  0.918 &  31.77 &  0.870 &  28.15 &  0.772 &  25.34 &  0.683 \\
KPN $\text{L}_{11}$ &  34.67 &  0.919 &  31.78 &  0.871 &  28.15 &  0.773 &  25.26 &  0.679 \\
KPN $\text{L}_{9}$  &  34.57 &  0.917 &  31.63 &  0.867 &  28.07 &  0.769 &  25.34 &  0.681 \\
KPN $\text{L}_{7}$  &  34.54 &  0.918 &  31.65 &  0.866 &  28.10 &  0.765 &  25.37 &  0.679 \\
KPN $\text{L}_{5}$  &  34.30 &  0.906 &  31.36 &  0.855 &  27.76 &  0.754 &  24.88 &  0.658 \\
KPN $\text{L}_{3}$  &  33.65 &  0.897 &  30.79 &  0.844 &  27.26 &  0.735 &  24.55 &  0.632 \\
KPN $\text{L}_{1}$  &  31.66 &  0.837 &  28.50 &  0.759 &  24.32 &  0.607 &  21.30 &  0.486 \\
\end{tabular}
}
\end{center}
\caption{Results on test datasets with different gains (noise levels) in measures of PSNR and SSIM, where Gain $\propto$ 8 is the noisiest. MKPN outperforms the state-of-the-art KPN~\cite{mildenhall2018burst} and other KPN variations.}
\label{table:denoising_exp}
\end{table}

Table~\ref{table:denoising_exp} compares the performance of different models 
and shows the average values for the four test datasets with different noise levels. It is to be noted that KPN~$\text{L}_5$ performs almost the same as KPN~\cite{mildenhall2018burst} with a reduced number of computations due to the separable kernels. Among the state-of-the-art models, we find that KPN~$\text{L}_{25}$ shows superior performance -- even better than KPN~\cite{mildenhall2018burst}. The results in Table~\ref{table:denoising_exp} also indicate that the results, in terms of PSNR and SSIM, get better as the size of the kernel increases. This can be attributed to the increased information that the kernels can extract from the surrounding region of the pixel and use it for better denoising. 
MKPN, on the other hand, combines the different kernels and outperforms all state-of-the-art models -- KPN~\cite{mildenhall2018burst} and the derived KPN~$\text{L}_n$ on all test datasets for each noise level. 

In Fig.~\ref{fig:example_bear} and Fig.~\ref{fig:example_grasshoper}, we inspect the denoised images from the test dataset to reveal the properties of different kernel sizes and the advantages of MKPN over KPN. 
The first row displays the full image while the second row shows an enlarged part marked by the green rectangle in the corresponding images in the first row.
The order of images in the columns is as follows -- noisy input burst, denoising results of KPN~\cite{mildenhall2018burst}, KPN~$\text{L}_{25}$, MKPN and the ground truth. 
In Fig.~\ref{fig:example_bear}, we find that the reconstructed image of KPN~\cite{mildenhall2018burst} is better than that of KPN~$\text{L}_{25}$. 
Although, in Table~\ref{table:denoising_exp}, KPN~$\text{L}_{25}$ performed superior to other state-of-the-art models, the denoised images from KPN~$\text{L}_{25}$ are oversmoothed due to the large size of the kernels. Hence, using a large kernel results in suboptimal visual quality of the denoised image. MKPN restores the structure of fur in Fig.~\ref{fig:example_bear} with more detail than other state-of-the-art models. 
In Fig.~\ref{fig:example_grasshoper},
KPN~\cite{mildenhall2018burst} fails to smooth out the background well enough. This is due to the small kernel trying to greedily denoise the local region, giving it an unrealistic texture. KPN~$\text{L}_{25}$, on the other hand, denoises the background efficiently, but fails to recover the sharp edges. With MKPN, the legs of the grasshopper are reconstructed  sharply while having a smooth background.

While KPN models with small kernels are able to reconstruct detailed features more successfully, their performance on flat, homogeneous areas is worse compared to the KPN models using larger kernels. On the other hand, KPN models with larger kernels tend to oversmooth the detailed features. 
MKPN manages to alleviate the shortcomings of the KPN models relying on fixed kernel sizes.

%% file: main.bbl
\begin{thebibliography}{10}

\bibitem{godard2018deep}
C.~Godard, K.~Matzen, and M.~Uyttendaele,
\newblock ``Deep burst denoising,''
\newblock in {\em European Conference on Computer Vision (ECCV)}, 2018, pp.
  538--554.

\bibitem{hdrplus}
S.~W. Hasinoff, D.~Sharlet, R.~Geiss, A.~Adams, J.~T. Barron, F.~Kainz,
  J.~Chen, and M.~Levoy,
\newblock ``Burst photography for high dynamic range and low-light imaging on
  mobile cameras,''
\newblock {\em ACM Transactions on Graphics (TOG)}, vol. 35, no. 6, 2016.

\bibitem{perona1990scale}
P.~Perona and J.~Malik,
\newblock ``Scale-space and edge detection using anisotropic diffusion,''
\newblock {\em IEEE Transactions on Pattern Analysis and Machine Intelligence},
  vol. 12, no. 7, pp. 629--639, 1990.

\bibitem{rudin1992nonlinear}
L.~I. Rudin, S.~Osher, and E.~Fatemi,
\newblock ``Nonlinear total variation based noise removal algorithms,''
\newblock {\em Physica D: Nonlinear Phenomena}, vol. 60, no. 1-4, pp. 259--268,
  1992.

\bibitem{burger2012image}
H.~C. Burger, C.~J. Schuler, and S.~Harmeling,
\newblock ``Image denoising: Can plain neural networks compete with bm3d?,''
\newblock in {\em IEEE Conference on Computer Vision and Pattern Recognition
  (CVPR)}, 2012, pp. 2392--2399.

\bibitem{jain2009natural}
V.~Jain and S.~Seung,
\newblock ``Natural image denoising with convolutional networks,''
\newblock in {\em Advances in Neural Information Processing Systems (NIPS)},
  2009, pp. 769--776.

\bibitem{vincent2010stacked}
P.~Vincent, H.~Larochelle, I.~Lajoie, Y.~Bengio, and P.-A. Manzagol,
\newblock ``Stacked denoising autoencoders: Learning useful representations in
  a deep network with a local denoising criterion,''
\newblock {\em Journal of Machine Learning Research}, vol. 11, pp. 3371--3408,
  2010.

\bibitem{xie2012image}
J.~Xie, L.~Xu, and E.~Chen,
\newblock ``Image denoising and inpainting with deep neural networks,''
\newblock in {\em Advances in Neural Information Processing Systems (NIPS)},
  2012, pp. 341--349.

\bibitem{agostinelli2013adaptive}
F.~Agostinelli, M.~R. Anderson, and H.~Lee,
\newblock ``Adaptive multi-column deep neural networks with application to
  robust image denoising,''
\newblock in {\em Advances in Neural Information Processing Systems (NIPS)},
  2013, pp. 1493--1501.

\bibitem{gondara2016medical}
L.~Gondara,
\newblock ``Medical image denoising using convolutional denoising
  autoencoders,''
\newblock in {\em IEEE International Conference on Data Mining Workshops
  (ICDMW)}, 2016, pp. 241--246.

\bibitem{resnet}
K.~He, X.~Zhang, S.~Ren, and J.~Sun,
\newblock ``Deep residual learning for image recognition,''
\newblock in {\em IEEE Conference on Computer Vision and Pattern Recognition
  (CVPR)}, 2016, pp. 770--778.

\bibitem{mao2016image}
X.~Mao, C.~Shen, and Y.-B. Yang,
\newblock ``Image restoration using very deep convolutional encoder-decoder
  networks with symmetric skip connections,''
\newblock in {\em Advances in Neural Information Processing Systems (NIPS)},
  2016, pp. 2802--2810.

\bibitem{brooks2018unprocessing}
T.~Brooks, B.~Mildenhall, T.~Xue, J.~Chen, D.~Sharlet, and J.~T. Barron,
\newblock ``Unprocessing images for learned raw denoising,''
\newblock {\em arXiv preprint arXiv:1811.11127}, 2018.

\bibitem{mildenhall2018burst}
B.~Mildenhall, J.~T. Barron, J.~Chen, D.~Sharlet, R.~Ng, and R.~Carroll,
\newblock ``Burst denoising with kernel prediction networks,''
\newblock in {\em IEEE Conference on Computer Vision and Pattern Recognition
  (CVPR)}, 2018, pp. 2502--2510.

\bibitem{liu2014fast}
Z.~Liu, L.~Yuan, X.~Tang, M.~Uyttendaele, and J.~Sun,
\newblock ``Fast burst images denoising,''
\newblock {\em ACM Transactions on Graphics (TOG)}, vol. 33, no. 6, pp. 232,
  2014.

\bibitem{Chen2018LearningTS}
C.~Chen, Q.~Chen, J.~Xu, and V.~Koltun,
\newblock ``Learning to see in the dark,''
\newblock in {\em IEEE Conference on Computer Vision and Pattern Recognition
  (CVPR)}, 2018.

\bibitem{kokkinos2018iterative}
F.~Kokkinos and S.~Lefkimmiatis,
\newblock ``Iterative residual cnns for burst photography applications,''
\newblock {\em arXiv preprint arXiv:1811.12197}, 2018.

\bibitem{frame_interpolation}
S.~Niklaus, L.~Mai, and F.~Liu,
\newblock ``Video frame interpolation via adaptive convolution,''
\newblock in {\em IEEE Conference on Computer Vision and Pattern Recognition
  (CVPR)}, 2017, vol.~1, p.~3.

\bibitem{Bako17}
S.~Bako, T.~Vogels, B.~McWilliams, M.~Meyer, J.~Nov\'ak, A.~Harvill, P.~Sen,
  T.~DeRose, and F.~Rousselle,
\newblock ``Kernel-predicting convolutional networks for denoising monte carlo
  renderings,''
\newblock {\em ACM Transactions on Graphics (TOG)}, vol. 36, no. 4, 2017.

\bibitem{vogels2018denoising}
T.~Vogels, F.~Rousselle, B.~Mcwilliams, G.~R{\"o}thlin, A.~Harvill, D.~Adler,
  M.~Meyer, and J.~Nov{\'a}k,
\newblock ``Denoising with kernel prediction and asymmetric loss functions,''
\newblock {\em ACM Transactions on Graphics (TOG)}, vol. 37, no. 4, pp. 124,
  2018.

\bibitem{mai2015kernel}
L.~Mai and F.~Liu,
\newblock ``Kernel fusion for better image deblurring,''
\newblock in {\em IEEE Conference on Computer Vision and Pattern Recognition
  (CVPR)}, 2015, pp. 371--380.

\bibitem{Rigamonti2013LearningSF}
R.~Rigamonti, A.~Sironi, V.~Lepetit, and P.~Fua,
\newblock ``Learning separable filters,''
\newblock {\em IEEE Transactions on Pattern Analysis and Machine Intelligence},
  vol. 37, pp. 94--106, 2013.

\bibitem{han2015deep}
S.~Han, H.~Mao, and W.~J. Dally,
\newblock ``Deep compression: Compressing deep neural networks with pruning,
  trained quantization and huffman coding,''
\newblock {\em arXiv preprint arXiv:1510.00149}, 2015.

\bibitem{WieArXiv18b}
S.~Wiedemann, A.~Marban, K.-R. M{\"u}ller, and W.~Samek,
\newblock ``Entropy-constrained training of deep neural networks,''
\newblock {\em arXiv preprint arXiv:1812.07520}, 2018.

\bibitem{WieArXiv18}
S.~Wiedemann, K.-R. M{\"u}ller, and W.~Samek,
\newblock ``Compact and computationally efficient representation of deep neural
  networks,''
\newblock {\em arXiv preprint arXiv:1805.10692}, 2018.

\bibitem{unet}
O.~Ronneberger, P.~Fischer, and T.~Brox,
\newblock ``U-net: Convolutional networks for biomedical image segmentation,''
\newblock in {\em International Conference on Medical Image Computing and
  Computer-Assisted Intervention (MICCAI)}, 2015, pp. 234--241.

\bibitem{open_images}
Google,
\newblock {\em Open Images Dataset V4},
\newblock Google, 2018,
\newblock \url{https://storage.googleapis.com/openimages/web/index.html}.

\bibitem{dng}
Adobe,
\newblock {\em Digital Negative (DNG) Specification},
\newblock Adobe, 2012,
\newblock
  \url{https://www.adobe.com/content/dam/acom/en/products/photoshop/pdfs/dng_spec_1.4.0.0.pdf}.

\bibitem{ssim}
Z.~Wang, A.~C. Bovik, H.~R. Sheikh, and E.~P. Simoncelli,
\newblock ``Image quality assessment: from error visibility to structural
  similarity,''
\newblock {\em IEEE Transactions on Image Processing}, vol. 13, no. 4, pp.
  600--612, 2004.

\end{thebibliography}
